\documentclass[11pt,a4paper]{article}
\pdfoutput=1
\usepackage[hyperref]{ranlp2021}
\usepackage{times}
\usepackage{latexsym}

\usepackage{microtype}

\aclfinalcopy % Uncomment this line for the final submission
\def\aclpaperid{245} %  Enter the acl Paper ID here

\usepackage{caption}
\usepackage{graphicx}
\usepackage[caption = false]{subfig}

\usepackage{listings}
\usepackage{multirow}
\usepackage{booktabs}
\usepackage{siunitx}

\usepackage{amsmath}

\title{System Combination for Grammatical Error Correction \\ Based on Integer Programming}

\author{Ruixi Lin \qquad Hwee Tou Ng \\
  Department of Computer Science \\ National University of Singapore \\
  \texttt{\{ruixi,nght\}@comp.nus.edu.sg} \\
}

\date{}

\begin{document}
\maketitle
\begin{abstract}
In this paper, we propose a system combination method for grammatical error correction (GEC), based on nonlinear integer programming (IP). Our method optimizes a novel $F$ score objective based on error types, and combines multiple end-to-end GEC systems. The proposed IP approach optimizes the selection of a single best system for each grammatical error type present in the data. Experiments of the IP approach on combining state-of-the-art standalone GEC systems show that the combined system outperforms all standalone systems. It improves \(F_{0.5}\) score by 3.61\% when combining the two best participating systems in the BEA 2019 shared task, and achieves \(F_{0.5}\) score of 73.08\%. We also perform experiments to compare our IP approach with another state-of-the-art system combination method for GEC, demonstrating IP's competitive combination capability.
\end{abstract}

\section{Introduction}
\label{intro}

Grammatical Error Correction (GEC) is the task of detecting and correcting grammatical errors of all types present in sentences in an essay, and generating a corrected essay~\cite{ng}.

Most of the latest GEC systems rely on pre-training with synthetic data and fine-tuning with task-specific data, and employ deep neural networks with attention mechanisms ~\cite{bryant}. Single GEC systems can be highly effective in capturing a wide range of grammatical error types ~\cite{ng}, but each individual system differs in its strengths and weaknesses in correcting certain error types, and the differences could result from the synthetic data used in pre-training a GEC system ~\cite{bryant}. Two main categories of synthetic data generation approaches have been introduced, including directly injecting noise into grammatically correct sentences according to error distributions~\cite{zhao,kakao}, or by back-translation ~\cite{sennrich,xie,tohoku}. While both categories could help a system achieve a high recall across many error types, it is hard to obtain a single uniform GEC system that is good at correcting all error types.

Presented with this difficulty and the strengths of individual systems, combining single GEC systems is thus a promising and efficient way to further improve precision and recall. Creating an ensemble of multiple systems is a common approach when it comes to combining multiple models, and the work of ~\citet{marcin} has shown its effectiveness when combining single GEC models with different random initializations and configurations. However, this mode of combination requires altering the component systems to achieve a tight integration.

In contrast, we focus on combination methods that need to consider only the outputs of individual systems. State-of-the-art system combination approaches working in this manner include ~\citet{raymond} and ~\citet{ibm}. The approach in ~\citet{raymond} adopts the MEMT system combination technique of ~\citet{memt} and learns a combined corrected sentence which is made up of different parts of multiple system outputs. The work by ~\citet{ibm} has proposed a system combination approach based on convex optimization. It treats single GEC systems as black boxes, rounding system weights to 0 or 1, and iteratively combines two systems at a time. 

In this paper, we propose a novel system combination method based on nonlinear integer programming (IP). Our method optimizes an $F$ score objective based on error types, and combines multiple component GEC systems simultaneously with binary selection variables. In Section \ref{sec:related}, we present related work on system combination for GEC. Then, our proposed IP approach is described in detail in Section \ref{sec:comb}. We present experimental setup and results in Section \ref{sec:exp}, provide analysis of results in Section \ref{sec:analysis}, and conclude in Section \ref{sec:conc}. Our source code is available at \url{https://github.com/nusnlp/gec_ip}.

\section{Related Work}
\label{sec:related}
System combination approaches for GEC are based on pipelining, confusion networks, error types, or optimization. Pipelining approaches, such as the CAMB system \cite{felice2014}, adopt a pipeline of simpler to more complex GEC systems for correction, but they suffer from error propagation. Confusion networks, especially the MEMT approach for GEC \cite{raymond}, learn the optimal word choice at a sentence location via a decoding scheme. The error type-based approaches aim at edit selection per error type and per system. The LAIX system \cite{laix} employs a confidence table and a rule-based conflict solver to select the optimal edits from component systems. Past work using integer linear programming (ILP) for GEC includes \cite{rozovskaya,yuanbin}. Moving on to using optimization-based selection variables, the IBM approach \cite{ibm} combines edits for a pair of systems at a time, based on error types and subsets of corrections. Continuous selection variables are learned by maximizing the subset-based $F_{0.5}$ objective. The IBM approach is most related to our proposed IP approach, and we highlight the key differences in Section \ref{subsec:keydiff}.

\section{System Combination}
\label{sec:comb}
\begin{figure*}
  \includegraphics[width=\textwidth]{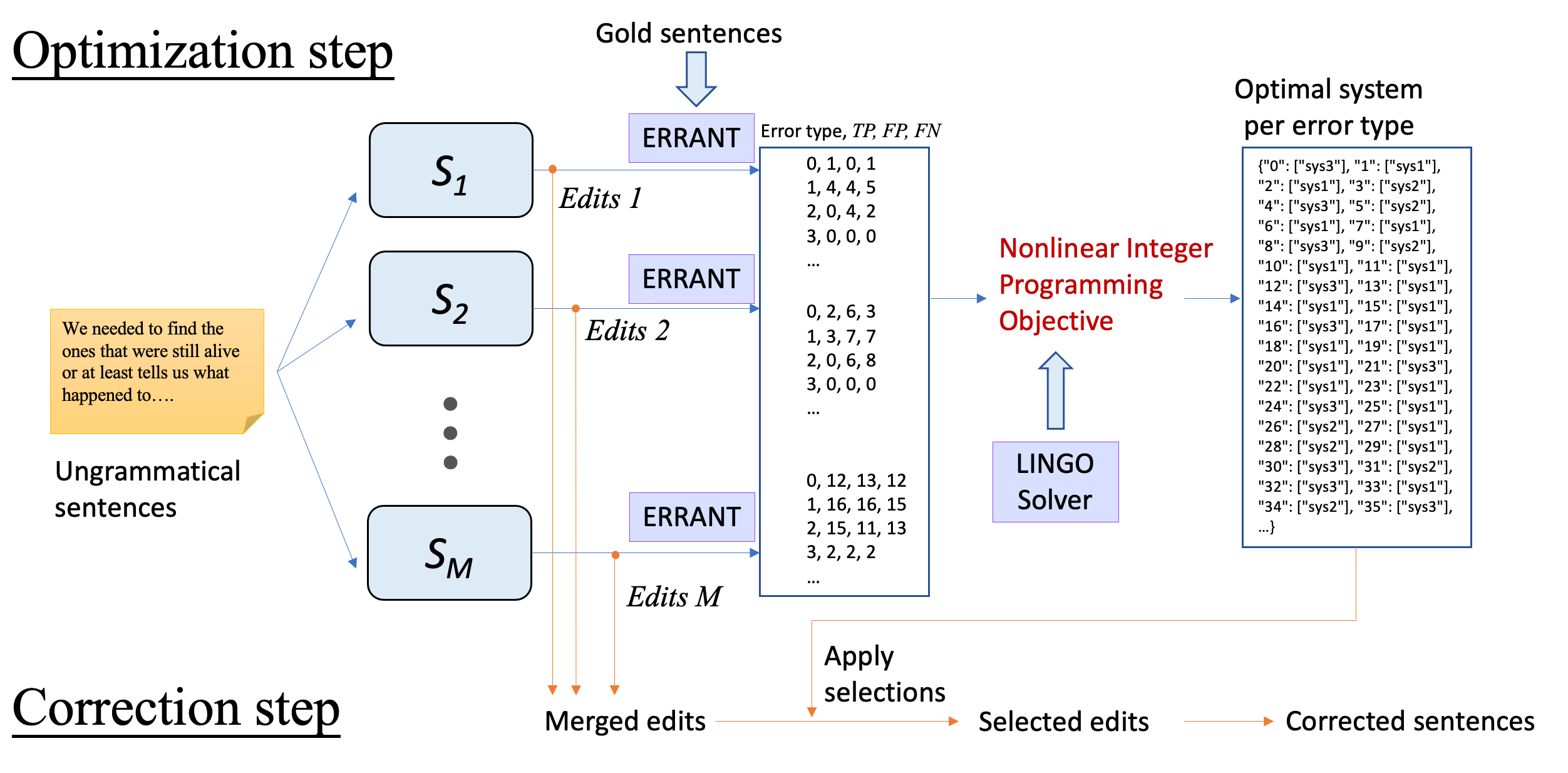}
  \caption{An overview of our proposed IP approach.}
  \label{fig:overview}
\end{figure*}

We combine systems based on the strengths of individual systems in terms of error types, and optimize directly the \textit{F} score evaluation metric \cite{fmeasure} to obtain error type-based selection variables for each system. Compared to ~\citet{ibm}, we make several major changes to achieve good precision and recall while making the combination more efficient. An overview of our proposed IP approach is illustrated in Figure  \ref{fig:overview}.

\subsection{An Integer Programming-Based Approach}
\label{subsec:ip}

First, we observe that the selection variables learned by convex optimization in ~\citet{ibm} are rounded to their nearest integers, either 0 or 1, for simplicity. This approximation of continuous variables raises the question of why binary variables were not directly used in the first place. A more direct solution is to adopt binary variables. Let
\begin{equation*}
x_{ij} = 
  \begin{cases}
    1 & \text{if $S_i$ is used to correct $T_j$} \\
    0 & \text{otherwise} \\
  \end{cases}
\end{equation*}
where $S_i$ refers to system $i$ in a set of $M$ systems  $S=\{S_1,\dots, S_M\}$ and $T_j$ refers to error type $j$ in a set of $N$ error types $T=\{T_1,\dots, T_N\}$. Taking $F$ score, the evaluation metric of GEC systems, as our objective function to maximize, we can formulate the GEC system combination problem as a nonlinear 0-1 integer programming (IP) problem as follows:
\begin{eqnarray*}
\lefteqn{\max F_{\alpha}(X) = } \\
& & \frac{(1+\alpha^2) \cdot TP_{sum}}{(1+\alpha^2) \cdot TP_{sum} + FP_{sum} + \alpha^2 \cdot FN_{sum}}
\end{eqnarray*}
s.t.
\begin{equation}\label{eq:ip5}
\sum\limits_{i \in S} x_{ij} = 1, \forall j \in T
\end{equation}
\begin{equation}\label{eq:ip8}
x_{ij} \in \{0, 1\}, \forall i \in S, j \in T
\end{equation}
where
\begin{equation}\label{ip2}
TP_{sum} = \sum\limits_{i \in S}\sum\limits_{j \in T}\lambda_{ij}^{TP}x_{ij}
\end{equation}

\begin{equation}\label{ip3}
FP_{sum} = \sum\limits_{i \in S}\sum\limits_{j \in T}\lambda_{ij}^{FP}x_{ij}
\end{equation}
\begin{equation}\label{ip4}
FN_{sum} = \sum\limits_{i \in S}\sum\limits_{j \in T}\lambda_{ij}^{FN}x_{ij}
\end{equation} 
Equation (\ref{eq:ip5}) imposes the constraint that each error type is corrected by exactly one system. Equation (\ref{eq:ip8}) is the integer constraint, resulting in a 0-1 integer programming model. In Equations (\ref{ip2}), (\ref{ip3}), and (\ref{ip4}), \(\lambda_{ij}^{TP}\), \(\lambda_{ij}^{FP}\), and \(\lambda_{ij}^{FN}\) respectively denote the true positive count, false positive count, and false negative count for system $i$ and error type $j$. In this paper, we set \(\alpha=0.5\) and optimize \(F_{0.5}\), the standard evaluation metric in GEC.

Moreover, the combination method in ~\citet{ibm} uses the intersection of the edits of multiple systems, which can be too sparse to be useful when many systems are combined. Their iterative combination approach may alleviate the sparsity problem, but high computational cost is incurred when the number of component systems is large, due to the inherent combinatorial explosion of finding the optimal order of combination. In contrast, our approach requires no subset splitting, and optimizes all component systems simultaneously as indicated in Equations (\ref{ip2}), (\ref{ip3}), and (\ref{ip4}). 

\subsection{Combination Procedure}
The input to the IP approach is $M$ corrected sentences (of the same ungrammatical sentence) given by GEC systems $S_1,\dots, S_M$. The output is a corrected sentence, after applying edit selection. The combination procedure consists of an optimization step followed by a correction (inference) step. Details are as follows.

\paragraph{Optimization Step.} We compute the true positive, false positive, and false negative counts for each error type and component system on a training dataset. Utilizing these counts, the 0-1 integer programming model defined in Section~\ref{subsec:ip} is solved using the commercial optimization software LINGO10.0\footnote{\url{https://www.lindo.com/index.php/products/lingo-and-optimization-modeling}} to compute the optimal solutions for \(x_{ij}\). In LINGO10.0, we adopt the INLP (integer nonlinear programming) model. For the experiments in this paper, the runtime for the LINGO solver to compute an optimal solution is 2 to 10 seconds.

\paragraph{Correction Step.} During correction (inference), the system applies an edit (by system $i$ to correct an error of type $j$) to an input sentence if \(x_{ij} = 1\) as determined by the LINGO solver. Conflicts of candidate edits from multiple systems (under different error types) can occur in the same location in a sentence. In other words, although a single system is used for each error type, a conflict can occur when two systems perceive an error in the same location to be of different error types, causing the location to have multiple candidate edits. When this happens, we set IP to randomly choose a candidate edit.

\subsection{Key Differences from the IBM approach}
\label{subsec:keydiff}
Since the IBM approach \cite{ibm} is the most related work, we summarize the key differences of our IP approach from the IBM approach.

\begin{enumerate}

    \item The IP approach directly combines all systems at once, as opposed to iteratively combining two systems at a time in the IBM approach, where the order of combination affects the outcome and there is a need to search for the best order of combination. In contrast, our approach avoids the problem of searching for the best order of combination.
    
    \item Binary (0-1) integer selection variables are directly used, in contrast to approximation by integers in the IBM approach.
    
    \item In the IP approach, we avoid having to perform subset splitting during optimization, in contrast to the IBM approach. For subset splitting, corrections from two systems are split into an intersection subset and subsets containing per system-only corrections, which can result in data sparsity.
    
\end{enumerate}

\section{Experiments}
\label{sec:exp}
\subsection{Component Systems}
\label{exp:individual}
To apply the IP combination method on GEC, we choose three state-of-the-art GEC systems as component systems. They are the systems UEdin-MS ~\cite{edin} and Kakao ~\cite{kakao} (the top two systems from the restricted track of the BEA 2019 shared task), as well as Tohoku ~\cite{tohoku}.

The three GEC systems share interesting commonalities and exhibit salient differences. The major commonalities are the use of the Transformer Big architecture ~\cite{transformers} (the Kakao system uses a variant, the copy-augmented Transformer ~\cite{zhao}), pre-training on 40M to 100M synthetic parallel data, ensemble of multiple models, and re-ranking. They differ in the synthetic data generation methods, the monolingual sources, the implementation of the architecture, and re-ranking features. These characteristics lead to individual strengths in correcting different types of errors, allowing room for improvements via system combination.

\subsection{Training and Evaluation}
\label{exp:data}

We use the official BEA 2019 shared task datasets for training and evaluation. We learn our selection variable values based on the output sentences of the component systems on the official validation set. At inference time, we apply combination on the output sentences of the component systems on the official blind test set of BEA 2019. The resulting output sentences are sent to the shared task leaderboard for evaluation, where \(P\), \(R\), and \(F_{0.5}\) scores are the evaluation metrics. The official scorer is the ERRANT evaluation toolkit v2.0.0 ~\cite{errant1}. 

\subsection{Experimental Results}
\label{exp:res}

Besides our proposed IP method, we compare with the MEMT-based system combination approach for GEC. We follow the work of ~\citet{raymond} to use the open source MEMT toolkit \cite{memt} for experiments.

MEMT system combination performs two major steps to combine edits: pairwise alignment and confusion network decoding with feature weight tuning. Pairwise alignment is first performed using METEOR \cite{meteor} to form a search space for combination. The alignment recognizes exact matches, words with the same stem, synonyms defined in WordNet, and unigram paraphrases. Then a confusion network is formed on top of the aligned sentences and beam search decoding is used to form hypotheses. During beam search, scoring of partial hypotheses is performed by a set of features, including hypothesis length to be normalized, log probability from a language model, n-gram backoff from the language model, and matched n-grams between the sentences generated by the component systems and the hypothesis. Tuning of feature weights is performed using ZMERT \cite{mert1}, optimized for the BLEU metric. We report the average score of three runs of each MEMT combination. 

The scores for the component systems and the IP combination approach are reported in Table \ref{res:component}, and the scores for the MEMT approach are reported in Table \ref{res:comb2}. All combination scores using the IP method are higher than the individual systems' scores. The best IP score is achieved by combining UEdin-MS and Kakao (1+2), and the \(F_{0.5}\) score is 73.08\%, which is 3.61\% higher than that of the best individual component system UEdin-MS. Comparing IP with MEMT, the average \(F_{0.5}\) score across all combinations of IP is 72.36\%, and that for MEMT is 71.42\%, so the average \(F_{0.5}\) score of IP is 0.94\% higher than that of MEMT. The IBM approach ~\cite{ibm} reported \(F_{0.5}\) score of 73.18\% when combining the component systems UEdin-MS and Kakao (1+2). Overall, the performance of IP is thus comparable to other state-of-the-art combination approaches.

 \begin{table}[t]
\begin{center}
\renewcommand\arraystretch{1.1}
\begin{tabular}{lccc}
 \toprule
 \textbf{System} & \(\textbf{P}\)    & \(\textbf{R}\)   & \textbf{F\textsubscript{0.5}} \\
 \hline
 1. UEdin-MS & 72.28 & 60.12 & 69.47 \\
 2. Kakao & 75.19 & 51.91 & 69.00 \\
 3. Tohoku & 74.71 & 56.67 & 70.24 \\
 \midrule
 \textbf{IP} &&&\\
 \hline
    C1: 1+2	    &\textbf{78.20}   &57.90 &\textbf{73.08}\\
    C2: 1+3	    &76.08	&\textbf{58.81}&71.86\\
  C3:  2+3	    &76.95	&55.54	&71.44\\
  C4:  1+2+3	&78.17	&57.88	&73.05\\
 \bottomrule
\end{tabular}
 \end{center}
 \caption{Scores (\%) of the component systems and the IP combination approach  on the BEA 2019 blind test set.}
 \label{res:component}
 \end{table}

\begin{table}[t]
\begin{center}
  \begin{tabular}{lSSSSSS}
    \toprule
    \multirow{2}{*}{\textbf{System}} &
      \multicolumn{3}{c}{\textbf{MEMT}} \\
      & {P} & {R} & {F\textsubscript{0.5}} \\
      \midrule
  C1: 1+2	    &72.52	&\textbf{60.92}	&69.90\\
    C2: 1+3		&73.06	&60.75	&70.29\\
  C3:  2+3	   	&75.84	&58.28	&71.50\\
  C4:  1+2+3    &\textbf{79.17}	&58.68	&\textbf{73.98}\\
    \bottomrule
  \end{tabular}
  \end{center}
  \caption{Scores (\%) of the MEMT approach on the BEA 2019 blind test set.}
 \label{res:comb2}
\end{table}

 \section{Analysis of Results}
\label{sec:analysis}

We analyze how much the combined system using the IP approach improves over individual component systems, on a per-sentence basis. Since reference edits are unavailable for the blind test set, we base our analysis on the BEA 2019 validation set and split it into two halves: the first half for training and the second half for testing. For each input sentence, we compare the \(F_{0.5}\) scores of its output sentences generated by Kakao, UEdin-MS, and the system obtained by IP combination of both. We assign each output sentence $s$ into one of two classes, based on whether (1) the \(F_{0.5}\) scores of $s$ are identical in Kakao and UEdin-MS; or (2) the \(F_{0.5}\) scores of $s$ are different. 

The findings are summarized as follows. Of the 2,192 test sentences, there are 1,503 sentences where Kakao and UEdin-MS have the same \(F_{0.5}\) score. For these sentences, the IP approach achieves the same or higher \(F_{0.5}\) score on 1,501 sentences. The \(F_{0.5}\) score of IP on these 1,501 sentences is 0.4\% higher than each individual system. For the remaining 689 sentences that either Kakao or UEdin-MS performs better, 503 out of 689 sentences benefit from IP combination, with an increase of 11.58\% in the overall \(F_{0.5}\) score on the 503 sentences compared to the average \(F_{0.5}\) score of Kakao and UEdin-MS. This analysis shows that the component systems benefit from the IP combination approach on a per-sentence basis.

\section{Conclusion}
\label{sec:conc}
In this paper, we have proposed a system combination approach for GEC based on nonlinear integer programming, which combines all systems at once. The use of binary selection variables is simpler and more direct, compared to using continuous variables then rounding them. The best $F_{0.5}$ score achieved is 73.08\% on the BEA 2019 test set.

\section*{Acknowledgements}

This research is supported by the National Research Foundation, Singapore under its AI Singapore Programme (AISG Award No: AISG-RP-2019-014). The computational work for this article was partially performed on resources of the National Supercomputing Centre, Singapore (https://www.nscc.sg).

\bibliographystyle{acl_natbib}
\bibliography{anthology,ranlp2021}

\end{document}